\documentclass[sigconf,nonacm]{acmart}
\usepackage{algorithm}
\usepackage{algorithmic}
\usepackage{booktabs}
\usepackage{subcaption}
\usepackage{placeins}
\AtBeginDocument{%
  }

\begin{document}


\title{Residual Vector-based Reconstruction as Long-Context Recall Regardless of Context Window Size
}
% Residual Activation Indexing and Query-Conditioned Gating

\author{MyungHoon Ryu}
% \authornote{}
\email{ryumh10@korea.ac.kr}
\affiliation{%
  \institution{Korea University}
  \country{South Korea}
}

\author{XinYu Piao}
% \correspondingauthor
\authornotemark[1]
\email{xypiao97@korea.ac.kr}
\affiliation{%
  \institution{Korea University}
  \country{South Korea}
}

\author{Jong-Kook Kim}
% \correspondingauthor
\email{jongkook@korea.ac.kr}
\affiliation{%
  \institution{Korea University}
  \country{South Korea}
}
\authornote{The authors are corresponding authors of this paper.}

\begin{abstract}
Large language models (LLMs) process long contexts, including long documents and lengthy conversations, but face token-level memory usage that increases proportionally to input length. Although model optimization and lossy prompt compression are widely used, these methods still fail to solve the long-context recall problem beyond pretrained and size-constrained context windows. This paper proposes a long-context recall method that maintains near-constant GPU memory usage as context length increases, without additional training.
The main idea is to reconstruct facts using parameter activations in the LLM’s feed-forward layers, which store residual vectors representing facts from the source document.
Utilizing residual vectors allows the LLM to deterministically reconstruct query relevant facts without referencing the original document, preserving high fidelity and reducing memory usage without fine-tuning weights.
Experimental results show that the proposed method enables answering single-fact questions in two-million-token story contexts where previous methods fail.

\end{abstract}


%% (teaser figure omitted for review submission)

\maketitle

\begin{figure*}[t]
\centering
\includegraphics[width=\textwidth]{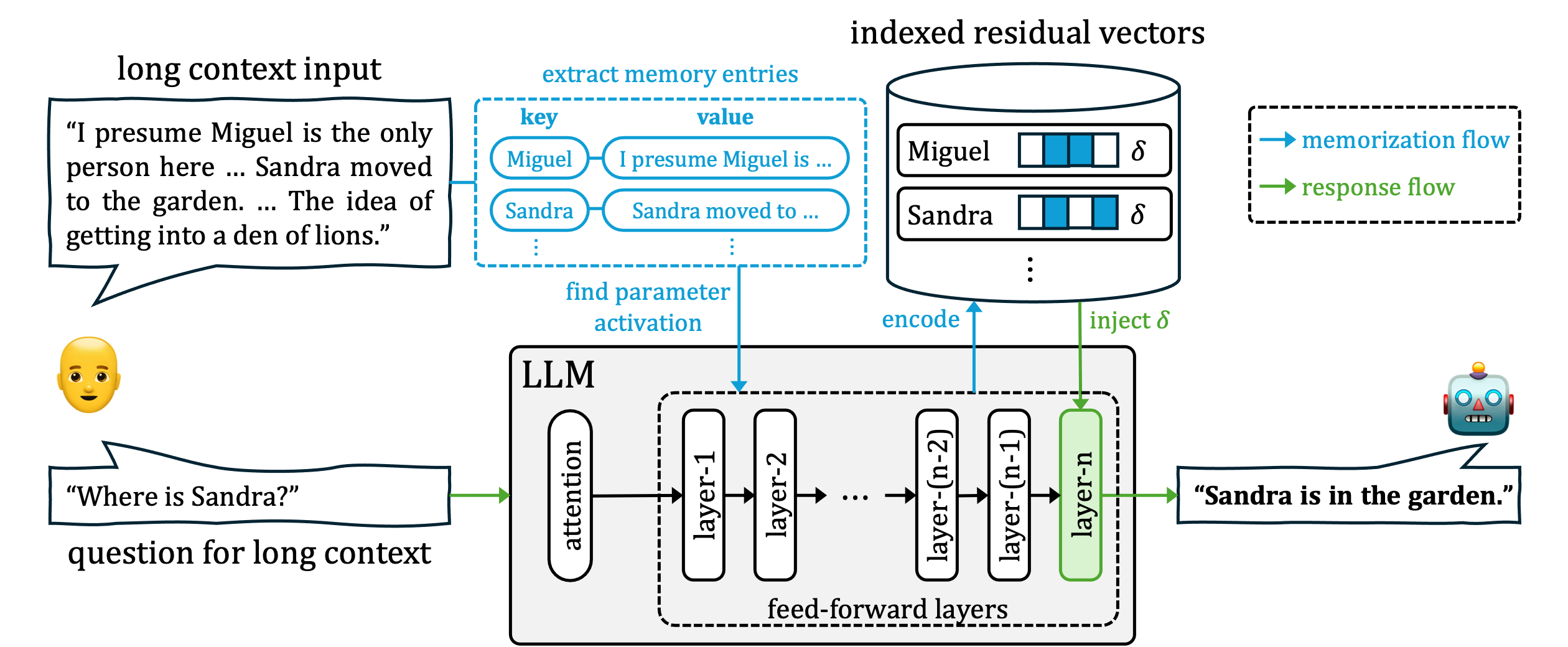}
\Description{Conceptual flow diagram. A blue memorization path extracts key and value entries from a document, finds their address activations in a frozen language model, encodes residual vectors, and stores the indexed vectors outside the model. A green response path routes a later question to memory, injects the selected residual vector at its feed-forward layer, and returns an answer.}
\caption{Overview of the proposed method in the two-phase flow. Memorization reads the source in bounded windows, extracts anchor key and value entries, encodes each value as a residual $\delta$ indexed by a model activation, and stores the indexed residual vector outside the frozen model before the source is discarded. At response time the question routes to stored entries, gated injection adds each selected $\delta$ at its assigned feed-forward layer, and the frozen model generates the answer from the reconstructed facts.}
\label{fig:overview}
\end{figure*}

\section{Introduction}

Large language models (LLMs) enable sustained interaction in document analysis, meeting assistance, literature review, legal reasoning, and multi-document synthesis~\cite{zhong2021qmsum,feng2021multidoc2dial,dasigi2021dataset,guha2023legalbench}. 
These interactions can span many turns, sessions, and source files~\cite{maharana2024evaluating,wu2025longmemeval}.
Later requests may require evidence, corrections, decisions, or intermediate conclusions introduced earlier in the interaction history~\cite{maharana2024evaluating,wu2025longmemeval}. 
Studies of conversational assistants report that users expect connectedness throughout a sequence of turns and treat common ground as remembered information~\cite{luger2016like,clark2019what,bickmore2005establishing}. 
Continuity behaviors in which an agent remembers earlier interactions and refers back to them increase liking and trust in long-term use~\cite{bickmore2005establishing}, whereas repeated assistant failures progressively erode user trust~\cite{luger2016like, clark2019what}.
Despite these user expectations, evaluations of long-term interaction show that LLMs lose relevant details and exhibit weak temporal or multi-session reasoning as dialogue histories grow~\cite{maharana2024evaluating,wu2025longmemeval}.
Although larger context windows increase available token capacity, token-level memory usage increases in proportion to input length.
Models still miss early or middle evidence and fail to connect distant facts~\cite{liu2024lost,hsieh2024ruler,zhang2024infty,kuratov2024babilong}. 
Long-context question answering requires reliable recall of evidence beyond the pretrained model's size-constrained context window~\cite{bai2024longbench,kuratov2024babilong,wu2025longmemeval}.

Documents longer than a pretrained context window cannot be processed by full-context inference as a single supported input. Training-based approaches extend this limit using positional scaling, long-context fine-tuning, or document-specific adaptation~\cite{chen2023extending,peng2024yarn,chen2024longlora,eyuboglu2025cartridges}. 
These approaches require additional optimization and an extended context window using positional scaling does not eliminate the inference-memory problem.
The key--value (KV) cache, which stores attention keys and values for processed tokens during generation, grows linearly as context length increases~\cite{xiao2024efficient,zhang2023h2o,li2024snapkv}. Training-free approaches instead reduce or externalize the retained context. Recurrent, streaming, cache-eviction, and prompt-compression methods discard or condense earlier tokens~\cite{dai2019transformer,bulatov2022recurrent,xiao2024efficient,zhang2023h2o,li2024snapkv,pan2024llmlingua2}.
These reductions can remove or distort context details required by a later question. Retrieval preserves the source but can omit required details from the selected answer context and its archive grows according to document length~\cite{lewis2020retrieval,kuratov2024babilong}. 
These limitations inevitably degrade the long-context question-answering performance of training-free methods beyond the pretrained context window.

This paper proposes a training-free method for selective factual recall regardless of the pretrained context window. 
Bounded input windows extract facts before future questions are known and encode each retained value as an activation-keyed residual vector outside the frozen language model.
Query-conditioned feed-forward gating reconstructs earlier factual statements after the source tokens and their KV cache have been discarded. 
A recalled fact decodes to the same stored sentence on every access and its storage cost is a single residual vector rather than token-level cache state.
A routing index of anchors and paraphrase questions resides outside the model to address the stored entries.
This method retains sentence-level details in long-context recall.
The evaluation covers three frozen model families and measures narrative and encyclopedic question answering, evidence-position sensitivity, reconstruction capacity, and GPU memory.
At two million tokens, single-fact accuracy reaches 50--80\% on all three models, while every compared training-free method falls to zero or near zero.
Stage-level analyses distinguish limitations arising from extraction, routing, reconstruction, and answer-model reasoning.
The main contributions are summarized as follows.
\begin{itemize}
    \item This paper proposes a two-phase training-free recall method that converts bounded document windows into query-agnostic fact records and reconstructs selected records after source removal, bounding GPU memory usage by a fixed fact budget rather than source length.
    \item The proposed method deterministically reconstructs source information while preserving high reconstruction fidelity and reduces recall storage by using activation-indexed parameter residual vectors, all without editing model weights.
    \item Experimental results show that the proposed method achieves 50--80\% single-fact accuracy at two million tokens and exceeds every evaluated training-free method on all three models, while its stored fact memory remains less than one gigabyte in every measured condition.
    \item A stage-level analysis distinguishes extraction, routing, reconstruction, and answer-model failures and identifies limited anchor correspondence as a central constraint on selective recall.
\end{itemize}

\section{Related Work}

Position scaling and long-context fine-tuning extend the supported input length after additional optimization, whereas document-specific adaptation internalizes a target document in model parameters~\cite{chen2023extending,peng2024yarn,chen2024longlora,eyuboglu2025cartridges}. 
Longer supported inputs still produce token-level inference state that grows according to input length.
Model editing adds or replaces explicitly specified factual associations in model parameters~\cite{meng2022locating,meng2022mass,fang2025alphaedit}.The resulting edits alter the knowledge available from the edited model, and the modification remains part of the weights. Truncation, recurrent processing, streaming inference, cache eviction, and cache selection restrict the active context or retained state, whereas prompt compression forms a shorter input before answer generation, and LLMLingua-2 implements task-agnostic compression as token classification~\cite{dai2019transformer,bulatov2022recurrent,xiao2024efficient,zhang2023h2o,li2024snapkv,pan2024llmlingua2}. All such query-agnostic reductions can remove evidence or sentence-level detail required by a question that arrives later. Retrieval-augmented generation retains the source text and an index, then selects relevant segments after a question arrives~\cite{lewis2020retrieval}. Detail loss occurs when a required segment does not enter the answer context rather than when the archive is formed. Its archive and index grow according to source size, and answer quality depends on retrieving every required segment.

Unlike these prior approaches, the proposed method optimizes only external residual vectors for selected facts while keeping model weights fixed. This avoids both the additional optimization of extension methods and the permanent modification of model editing. Because the underlying model remains unchanged, removing this external store completely restores the baseline model to its original state. Questions are answered without the original document by retaining sentence-level fact records, which are selected before future queries arrive and encoded under a fixed budget.

\section{Proposed Method}
\label{sec:method}

\subsection{Overview}

\begin{figure*}[t]
\centering
\includegraphics[width=\textwidth]{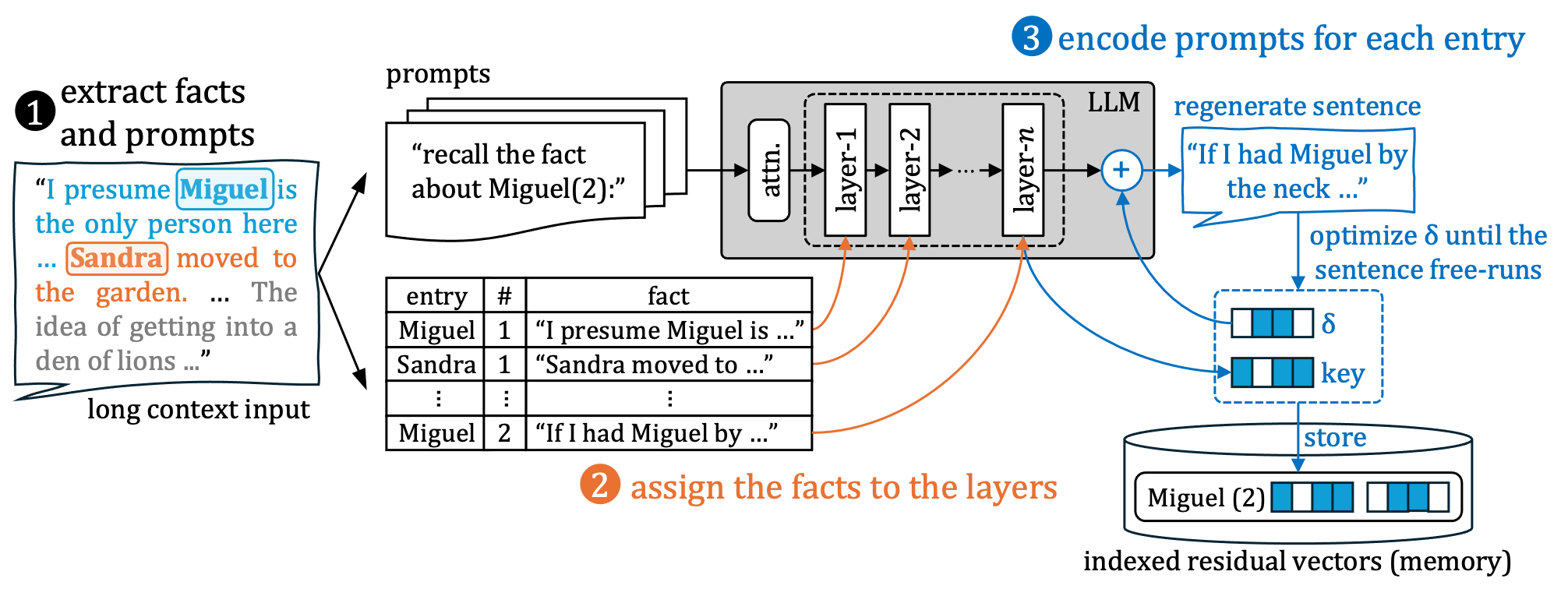}
\Description{Three-stage memorization flow. Stage 1 extracts Miguel and Sandra facts from a source document and lists each anchor, fact index, and sentence as a fact record. Stage 2 assigns the records to feed-forward layers. Stage 3 forms an indexed recall prompt for each record, captures its activation key at the assigned layer, optimizes residual delta until the sentence free-runs, and stores the key and residual vector outside the model.}
\caption{The memorization phase in three stages. Stage 1 extracts query-agnostic fact records from the source and forms indexed anchor and value entries, and a fixed capacity budget caps the retained count. Stage 2 assigns each entry to a storage layer in round-robin order. Stage 3 replays an entry-specific recall prompt, records the activation key at the assigned layer, and optimizes the residual vector until the stored sentence can be regenerated. The entries and their routing index persist after the source tokens are discarded.}
\label{fig:memorization}
\end{figure*}

\begin{figure*}[t]
\centering
\includegraphics[width=\textwidth]{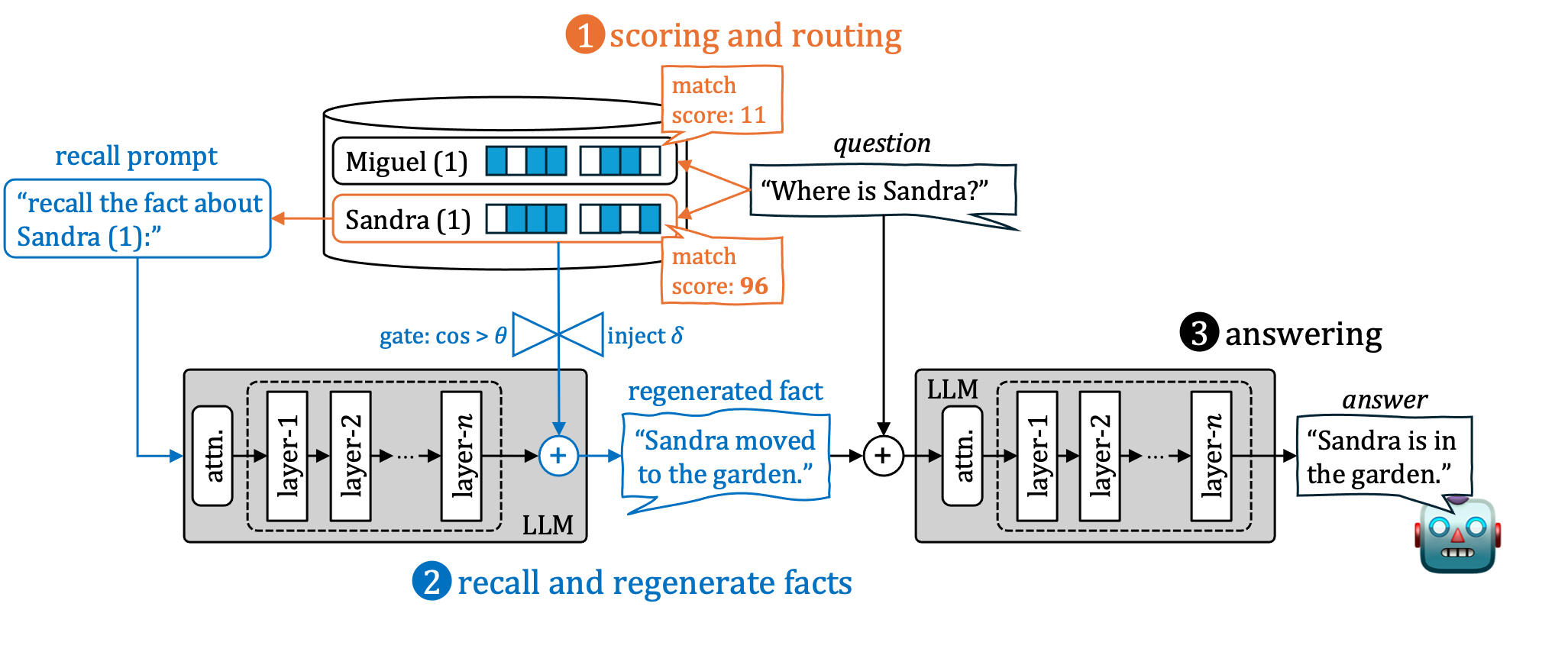}
\Description{Three-stage response flow. Stage 1 scores the stored Miguel and Sandra entries for the question Where is Sandra and selects Sandra. Stage 2 forms the recall prompt for Sandra 1, applies a cosine gate, injects the selected residual vector at its assigned layer, and regenerates the sentence Sandra moved to the garden. Stage 3 supplies the regenerated fact and question to the frozen model, which answers that Sandra is in the garden.}
\caption{The response phase in three stages. Stage 1 scores stored anchors against the question and selects the Sandra anchor after reciprocal-rank fusion. The selected anchor expands to Sandra (1). Stage 2 replays the entry-specific recall prompt, and the similarity gate injects the matching residual vector at its assigned layer to reconstruct the stored sentence. Stage 3 places reconstructed facts in source order in the answer prompt, and the frozen model generates the final answer.}
\label{fig:response}
\end{figure*}

Figure~\ref{fig:overview} separates the method into memorization and response phases.
The language model remains frozen in both phases.
The model diagram depicts every feed-forward layer, while memory storage uses a selected subset of these layers.
The illustrated document provides facts anchored by Miguel and Sandra. 
During memorization, each anchor and fact index defines a recall key. 
A fixed recall prompt binds that key to a parameter activation, and per-fact optimization encodes the sentence value into the residual $\delta$ shown in the figure.
The activation address and residual remain in an external indexed store.
The response phase begins after removal of the source tokens and their KV cache.
The Sandra question routes to the corresponding entry. 
Gated residual-vector injection reconstructs the sentence stating that Sandra moved to the garden, and a separate generation pass produces the output answer.
Retained state follows the fact budget rather than source length.
The formal notation for these components appears in the two phase subsections.
Figures~\ref{fig:memorization} and~\ref{fig:response} detail the two phases.

\subsection{Memorization Phase}
\label{sec:memorization}

\paragraph{Fact extraction.}
The memorization phase converts document $D$ into memory $\mathcal{M}$ and routing index $\mathcal{I}$. Stage 1 in Figure~\ref{fig:memorization} begins by selecting candidate sentences that carry factual signals. A high-recall candidate filter scans segmented sentences in one pass and retains sentences that contain entities, numerals, or identifier-like tokens. The frozen language model reads retained candidates in fixed-size batches, which bounds memorization-time working memory for any document length. The model proposes a short verbatim anchor for each candidate. A deterministic sentence span replaces an invalid or declined proposal. Each accepted candidate creates a fact record $(e,j,v,o)$. The table in Figure~\ref{fig:memorization} visualizes the anchor $e$, fact index $j$, and verbatim sentence value $v$. The example shows two facts assigned to the Miguel anchor and one fact assigned to the Sandra anchor. Indices 1 and 2 distinguish the two Miguel values. Anchor $e$ serves as the routing target, index $j$ distinguishes facts sharing the same anchor, and value $v$ defines the recall target. Source offset $o$ restores document order. Extraction remains query agnostic because no question exists during memorization. Key $\kappa(e,j)$ combines the anchor and fact index. The appended index creates distinct recall prompts and separable key activations for facts sharing the same anchor. Capacity allocation then determines which records proceed to layer assignment.

\paragraph{Capacity allocation.}
Storage budget $B$ governs retention before Stage 2 in Figure~\ref{fig:memorization} and denotes the maximum number of retained facts. The budget is
\begin{equation}
B \;=\; \min\!\left(\hat{m}\, n_{\mathrm{anc}},\; K C\right).
\label{eq:budget}
\end{equation}
The two arms of the minimum impose information and capacity bounds. The information arm $\hat{m}\,n_{\mathrm{anc}}$ scales according to the number of extracted anchors $n_{\mathrm{anc}}$, where retention depth $\hat{m}$ caps the facts retained per anchor. The capacity arm $KC$ caps total allocation, where $K$ is the number of storage layers and $C$ is the nominal allocation limit per layer. Parameter $C$ does not guarantee reconstruction capacity for every fact distribution. If $n_{\mathrm{anc}}\leq B$, each anchor may retain up to $m=\lfloor B/n_{\mathrm{anc}}\rfloor$ latest facts. Later statements update earlier entity states in the BABILong state-tracking tasks, which motivates latest-first retention. If $n_{\mathrm{anc}}>B$, anchors are ranked first by the number of model-validated facts and then by total retained facts. The leading $B$ anchors retain their latest fact. This second case occurs in some of the longest encyclopedic conditions. Retained count $N$ never exceeds $B$, and an empty extraction set produces empty memory. Before discarding the source, memorization constructs one routing document per retained anchor from the anchor, retained values, and paraphrase questions generated from those values. 
Dense and lexical representations rank anchors at query time. Their routing documents select eligible entries, while the residual vectors reconstruct the values included in the answer prompt. Recall reads each value from a single activation address, which permits entry-level deletion of the stored representation.

\paragraph{Layer assignment.}
Let $n_{\mathrm{layer}}$ denote the total number of feed-forward layers in the frozen language model $f$, presented as $n$ in the figures. A held-out recall probe measures reconstruction quality after residual-vector injection at each candidate layer and selects $K$ storage layers from the full set $\{1,\ldots,n_{\mathrm{layer}}\}$. The selected layers form the ordered tuple $\mathcal{L}=(\lambda_0,\ldots,\lambda_{K-1})$. Stage 2 in Figure~\ref{fig:memorization} distributes the retained entries to this selected set. The orange arrows connect individual rows to assigned members of $\mathcal{L}$ rather than every layer in $f$. Storing every fact at one layer would make all entries compete inside a single associative store. Let $\iota(e)$ be the zero-based extraction-order index of anchor $e$ and let $j$ be the zero-based fact position inside that anchor. Fact $j$ is assigned according to
\begin{equation}
\ell(e,j) \;=\; \lambda_{(\iota(e)+j)\bmod K} .
\label{eq:layer}
\end{equation}
The offset $\iota(e)$ staggers the starting layer between anchors, and the round-robin schedule spreads entries evenly onto the selected layers and prevents capacity saturation at any single layer.

\paragraph{Memory encoding.}
Stage 3 of Figure~\ref{fig:memorization} converts each layer-assigned record into the indexed residual-vector memory shown at the lower right. The stacked sheets denote one recall prompt per retained entry. The prompt shown for Miguel (2) produces an activation key at its assigned layer. Residual $\delta$ is optimized until the associated sentence can be regenerated. Transformer feed-forward layers have been characterized as key--value memories~\cite{geva2021transformer}. An activation inside this layer can serve as an address. A fixed recall prompt $\rho(\kappa)$ asks the model to reproduce the fact identified by key $\kappa$. Its fixed wording makes the same address reappear at response time. The evaluated template is \texttt{Recall the fact about \{key\}:}. The recall key is the activation $a^{(\ell)}(\kappa)\in\mathbb{R}^{d_{\mathrm{ff}}}$ captured at the input of down projection $W_d^{(\ell)}\in\mathbb{R}^{d_{\mathrm{model}}\times d_{\mathrm{ff}}}$ at the final key token of the prompt. Width $d_{\mathrm{ff}}$ is the feed-forward hidden dimension, and $d_{\mathrm{model}}$ is the model dimension.
Candidate residual $\delta\in\mathbb{R}^{d_{\mathrm{model}}}$ represents the content added during recall. Memory encoding minimizes
\begin{equation}
\mathcal{L}_{\mathrm{enc}}(\delta;\kappa,v,u)
=-\sum_{t=1}^{|v|}\log p_{f\oplus_{\ell}\delta}
\!\left(v_t\,\middle|\,\rho(\kappa),u_{<t}\right).
\label{eq:encode}
\end{equation}
Eq.~\eqref{eq:encode} sums the negative log-likelihood of value tokens $v_t$. Prefix $u_{<t}$ separates two optimization regimes. Distribution $p_{f\oplus_{\ell}\delta}$ follows residual-vector injection at layer $\ell$ of frozen model $f$. During encoding, injection occurs only at the recall-key token position that carries the address activation. Teacher forcing supplies true prefix $u_{<t}=v_{<t}$. After 40 guided steps, free-running decoding tests autoregressive reproduction of $v$. A failed fact receives 20 refinement steps using model-generated prefix $u_{<t}=\hat{v}_{<t}^{(\delta)}$. Per-fact encoding starts from $\delta=0$ and optimizes only $\delta$. Model parameters remain fixed, and the optimized vector becomes residual vector $r$. The stored key is the normalized captured activation $\hat{k}=a^{(\ell)}(\kappa)/\lVert a^{(\ell)}(\kappa)\rVert$, and a retained entry satisfies $a^{(\ell)}(\kappa)\neq 0$. The memory stores $r$ beside $\hat{k}$. After encoding $N$ facts, the memory is
\[
\mathcal{M}=\bigl\{(\kappa_i,\hat{k}_i,r_i,\ell_i,o_i,t_i) \mid 1\leq i\leq N\bigr\}.
\]
The value $N$ denotes the number of retained facts. Each $\ell_i$ belongs to $\mathcal{L}$. Source offset $o_i$ restores order, and token length $t_i$ bounds decoding. Symbol $\kappa_i$ reconstructs the fixed recall prompt. Keys and residual vectors remain external to the model. The processed bounded input window is no longer required after encoding. Removal of $\mathcal{M}$ restores the original inference path. Storage requires $N(d_{\mathrm{ff}}+d_{\mathrm{model}})$ scalars plus indexed prompt keys, layer identifiers, source offsets, value token lengths, and the routing index.

\subsection{Response Phase}
\label{sec:response}

\paragraph{Query routing.}
Stage 1 in Figure~\ref{fig:response} uses question $q$ to score and select anchors from $\mathcal{I}$. In the illustrated example, the question assigns an illustrative match score of 96 to the Sandra anchor and 11 to the Miguel anchor. Sandra (1) defines the recall prompt shown in the figure before any fact is reconstructed. Dense semantic ranking handles paraphrased questions, and lexical ranking based on inverse-document-frequency weights preserves rare names and identifiers. Reciprocal rank fusion combines the two rankings~\cite{cormack2009reciprocal}. A verbatim anchor in the question restricts routing to exact matches. In other cases, a candidate must reach 0.6 times the leading fused score. Each selected anchor expands to indexed fact keys that carry separate recall prompts $\rho(\kappa)$. Routing selects at most five anchors, and recall decodes at most 24 facts in fused-rank order. Each routed prompt must still win the gated similarity competition in Stage 2 of Figure~\ref{fig:response}.

\paragraph{Gated recall.}
Stage 2 in Figure~\ref{fig:response} presents a recall prompt for each routed fact and regenerates its stored sentence. 
Because decoding is greedy, a routed fact reproduces the same output at every recall.
Facts stored at layer $\ell$ form an associative memory~\cite{ramsauer2020hopfield}. Let $\mathcal{S}_{\ell}=\{i\mid \ell_i=\ell\}$ denote the indices of all entries stored at layer $\ell$. A read hook installed at the assigned layer compares prompt activation $x\in\mathbb{R}^{d_{\mathrm{ff}}}$ against every stored key $\hat{k}_i\in\mathbb{R}^{d_{\mathrm{ff}}}$ in $\mathcal{S}_{\ell}$. The hook observes the activation at every token position and every decoding step, and the gate shown in Stage 2 decides when injection happens. The plus symbol is placed after the layer stack for visual clarity but denotes addition at the assigned layer output.
\begin{subequations}
\label{eq:read}
\begin{equation}
s_i(x) = \cos\!\left(x,\hat{k}_i\right),
\label{eq:read-sim}
\end{equation}
\begin{equation}
\alpha_i(x) =
\frac{\exp\!\left(\beta s_i(x)\right)}
{\sum_{j\in\mathcal{S}_{\ell}} \exp\!\left(\beta s_j(x)\right)},
\label{eq:read-weight}
\end{equation}
\begin{equation}
g(x) = \mathbb{I}\!\left\{
\max_{i\in\mathcal{S}_{\ell}} s_i(x)>\theta\right\},
\label{eq:read-gate}
\end{equation}
\begin{equation}
\widetilde{y}^{(\ell)}(x)
= W_d^{(\ell)}x
+ g(x)\sum_{i\in\mathcal{S}_{\ell}} \alpha_i(x)r_i.
\label{eq:read-out}
\end{equation}
\end{subequations}
The gate of Eq.~\eqref{eq:read-gate} opens when the strongest similarity exceeds threshold $\theta\in[-1,1]$. Symbol $\mathbb{I}\{\cdot\}$ denotes the indicator function. Inverse temperature $\beta>0$ concentrates Eq.~\eqref{eq:read-weight} on the best-matching key. Eq.~\eqref{eq:read-out} adds the weighted residual vectors to the layer output. Modified output $\widetilde{y}^{(\ell)}(x)$ and residual $r_i$ belong to $\mathbb{R}^{d_{\mathrm{model}}}$. Each recall prompt installs the gated read only at its assigned layer and confines competition to entries stored at that layer. The gate icon in Stage 2 of Figure~\ref{fig:response} represents the cosine condition at the evaluated threshold $\theta{=}0.6$. Nonwinning softmax weights remain nonzero, while $\beta{=}50$ sharpens the distribution toward the leading key. Because the implementation applies no explicit position mask, the similarity condition governs every observed token position.
Encoding fits each residual vector at the single recall-key activation, while response-time gating evaluates the cosine condition at every prefill and decoding position and injects only where it holds. The high inverse temperature makes the read effectively select the top-1 key, and a hard top-1 rule is a direct substitute.

\paragraph{Answer generation.}
Stage 3 in Figure~\ref{fig:response} supplies the regenerated facts and original question $q$ to a standard generation pass. Each stored value is reconstructed independently, and decoding stops at the token length recorded during encoding. Source offsets order the facts, and anchor tags identify them in the answer prompt. Answer-context length depends on the routed values rather than the complete source.

\begin{figure*}[t]
\centering
\begin{subfigure}{\textwidth}
\centering
\includegraphics[width=0.99\textwidth]{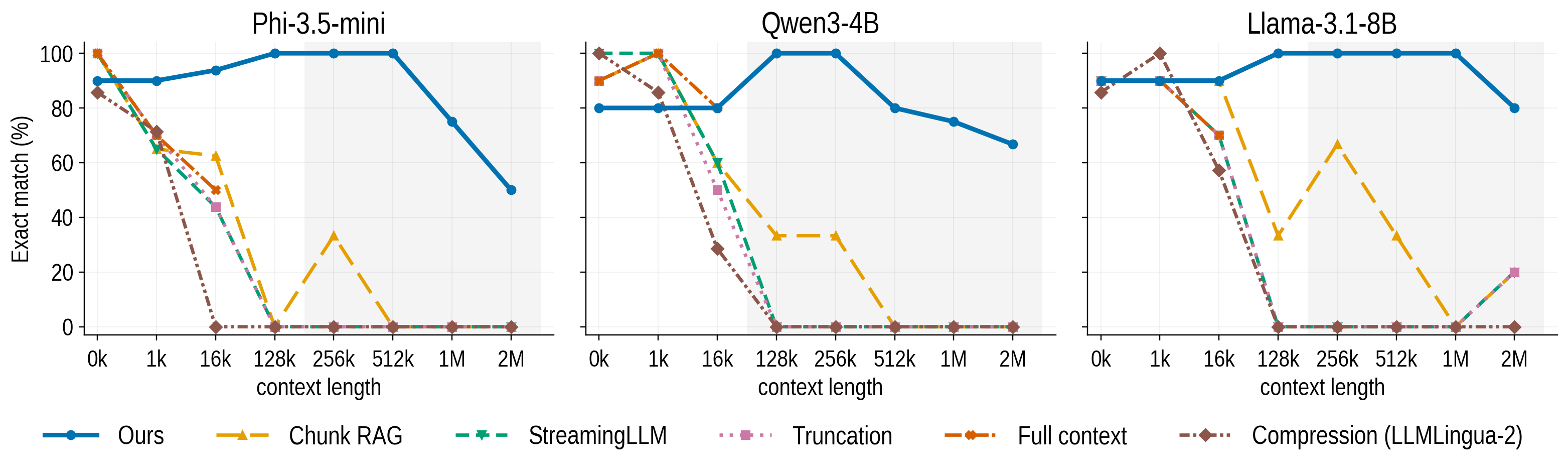}
\Description{Line charts for three models showing QA1 accuracy from zero to two million tokens. Token-budget methods approach zero at long inputs, Chunk RAG is uneven there, and the proposed method stays between fifty and one hundred percent.}
\caption{Single-fact recall from 0K to 2M.}
\label{fig:qa1_length_sweep}
\end{subfigure}
\begin{subfigure}{\textwidth}
\centering
\includegraphics[width=0.99\textwidth]{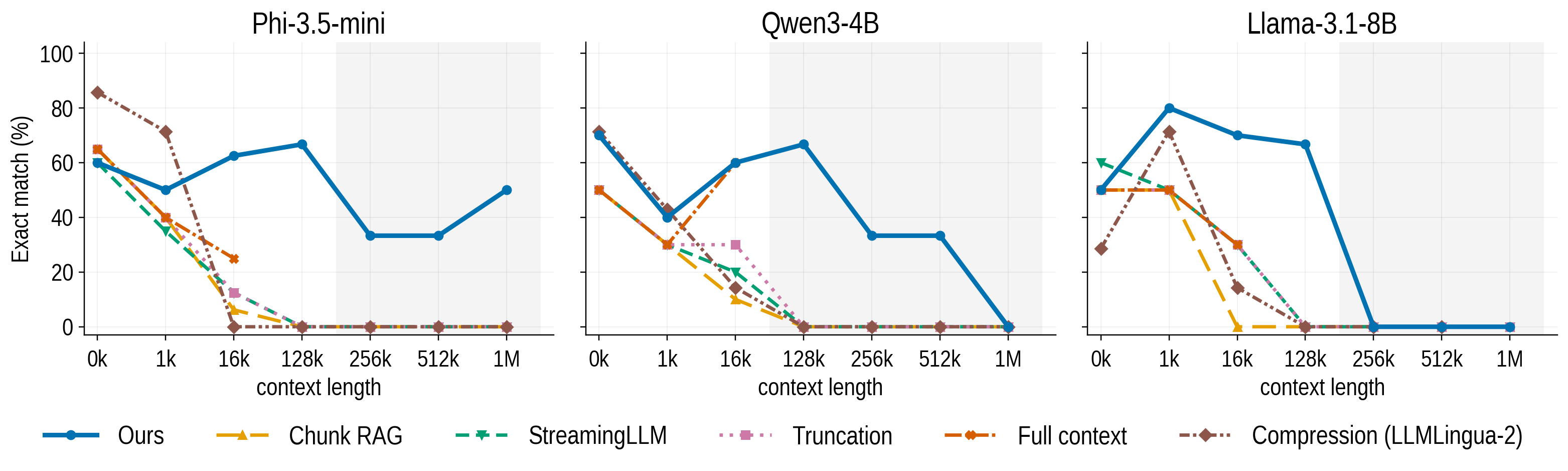}
\Description{Line charts for three models showing QA2 accuracy from zero to one million tokens for the same methods as the QA1 panel.}
\caption{Two-fact composition up to the 1M boundary.}
\label{fig:qa2_length_sweep}
\end{subfigure}
\caption{BABILong accuracy on one shared length axis. The shaded region lies beyond each model's pretrained context window. The QA1 panel ends at 2M, and the QA2 panel ends at 1M.}
\label{fig:length_sweeps}
\end{figure*}

\section{Evaluation}
\label{sec:setup}

\begin{table}[tb]
\centering
\caption{Proposed-method sample counts in the length comparisons. Each entry lists Phi-3.5-mini, Qwen3-4B, and Llama-3.1-8B in that order. A dash marks an unevaluated condition. The task heatmap pools additional samples in some cells. RULER compression and closed-book scores each use eight questions per model and length. Their sample counts differ from the proposed method at longer inputs.}
\label{tab:samples}
\begin{tabular}{@{}lccc@{}}
\toprule
Length & BABILong QA1 & BABILong QA2 & RULER QA \\
\midrule
0K & 20/10/10 & 20/10/10 & -- \\
1K & 20/10/10 & 20/10/10 & -- \\
16K & 16/10/10 & 16/10/10 & 8/8/8 \\
128K & 3/3/3 & 3/3/3 & 2/3/2 \\
256K & 3/3/3 & 3/3/3 & -- \\
512K & 5/5/3 & 3/3/3 & 2/3/2 \\
1M & 4/4/4 & 2/1/1 & 1/3/1 \\
2M & 4/3/5 & --/--/-- & 1/2/1 \\
\bottomrule
\end{tabular}
\end{table}

\subsection{Experimental Setup}

\paragraph{Language models}
Experiments evaluate three frozen instruction-tuned models. These models are Phi-3.5-mini, Qwen3-4B~\cite{yang2025qwen3}, and Llama-3.1-8B~\cite{grattafiori2024llama}. Phi-3.5-mini follows the Phi-3 family~\cite{abdin2024phi3technicalreporthighly} and contains 32 multi-head attention layers. Qwen3-4B and Llama-3.1-8B use grouped-query attention and contain 36 and 32 layers. Phi-3.5-mini has 3.8B parameters, a context window of 131{,}072 tokens, hidden width 3{,}072, and feed-forward width 8{,}192. Qwen3-4B has 4.0B parameters, a 40{,}960-token window, hidden width 2{,}560, and feed-forward width 9{,}728. Llama-3.1-8B has 8.0B parameters, a 131{,}072-token window, hidden width 4{,}096, and feed-forward width 14{,}336. Context-window size determines the supported full-context range. Feed-forward width determines the activation-key dimension of each layer-specific store. The three configurations test the method in distinct context limits and activation spaces.

\paragraph{Implementation}
Forward passes use bf16 numerics, and residual-vector optimization uses fp32 numerics.
Experiments run on one 80 GB A100 GPU. 
The benchmark harness limits full-context QA evaluation to inputs no longer than 40{,}000 tokens.
Anchor assignment reads 12 candidates per batch. 
A development recall probe ranks candidate layers by exact reconstruction, and the default configuration uses the leading $K{=}12$ probe-validated layers for all three models.
The gate threshold is $\theta{=}0.6$, and inverse temperature $\beta{=}50$ concentrates the read on one key.
Retention depth is $\hat{m}{=}14$ in the main evaluations. 
BGE-M3~\cite{chen2024bge} encodes the routing documents for the proposed method and the text chunks for Chunk RAG.
Gate threshold, read sharpness, retention depth, routing limits, and optimization schedule remain fixed outside the stated ablations. 
Settings specific to one experiment, namely the per-layer allocation $C$, the storage-layer count at the longest evaluated lengths, and the routing retention depth, appear where that experiment is reported.

\paragraph{Benchmarks}
BABILong~\cite{kuratov2024babilong} tests factual recall and composition by embedding synthetic facts in book-length text. Evaluation covers QA1 and QA2 length sweeps and a QA1 to QA5 task grid. RULER QA~\cite{hsieh2024ruler} tests recall in encyclopedic prose by placing SQuAD evidence~\cite{rajpurkar2016squad} in distractor passages. The ablations examine reconstruction capacity, evidence position, and routing, and the routing probe derives from NoLiMa~\cite{modarressi2025nolima}.
Development and evaluation samples are disjoint. BABILong development uses the first 80 samples of each configuration, and the main evaluation draws from samples 80 to 160. Official splits at 256K and longer are reserved for evaluation. BABILong and RULER inputs originate from the official generators, and the NoLiMa probe uses the official needle sets and haystacks. At the longest evaluated lengths, deterministic anchor rules replace language-model anchor refinement. Memorization receives no evaluation question or answer. Table~\ref{tab:samples} lists the proposed method's sample counts for the length comparisons.

\paragraph{Compared methods}
Truncation retains the latest 4{,}096 tokens. The official StreamingLLM implementation~\cite{xiao2024efficient} targets continuous autoregressive streams and requires attention and positional cache operations specific to each model. Its original release lacks a unified path for the three evaluated model configurations. 
The present benchmark provides a completed document before question answering rather than a continuous generation stream.
The evaluated prompt proxy uses a 4{,}096-token budget and retains four initial attention sink tokens plus the most recent 4{,}092 tokens. It represents the initial-token and recent-token retention policy rather than the complete rolling KV cache implementation. Figures and tables use the short label StreamingLLM for this proxy. Chunk RAG~\cite{lewis2020retrieval} uses BGE-M3~\cite{chen2024bge} to index non-overlapping 384-token chunks. It retrieves 20 chunks, reranks them using a BGE cross-encoder, and places the leading three in document order in the answer prompt. Retrieval budgets remain fixed for comparability. The LLMLingua-2~\cite{pan2024llmlingua2} method compresses the source, independent of the question, to the same 4{,}096-token cap before inference. Full context serves as an in-window reference at inputs no longer than the 40{,}000-token harness limit. A closed-book condition estimates answers attributable to pretraining knowledge on encyclopedic benchmarks.

\paragraph{Metrics}
Every method receives the same instruction provided by the BABILong harness and inherited by the RULER runs.
Answer scoring follows each benchmark's scoring procedure. The primary accuracy metric is exact match (EM), the percentage of evaluated questions whose generated answer is judged correct by the benchmark string check. RULER QA uses \texttt{string\_match\_part}, which marks an answer correct when a gold alias occurs as a substring of the generated answer. BABILong and NoLiMa use a substring exact-match check against the gold answer. Stage-level metrics locate failures before answer generation. Extraction coverage records whether the retained fact set contains the gold evidence. Routing recall records whether the target anchor enters gated recall. Reconstruction fidelity is the token-level F1 between a decoded value and its stored sentence. Table~\ref{tab:samples} records sample count $n$ for the proposed method. Resource measures distinguish activation-store size from peak GPU memory.

\subsection{Long-Context Recall and Composition}
This experiment measures single-fact recall and two-fact composition on BABILong from 0K to 2M lengths. 
It uses the narrative per-layer allocation  $C{=}4096$, and three cells at the longest lengths raise the storage-layer count to $K{=}14$, namely Phi-3.5-mini at 1M and 2M and Qwen3-4B at 2M. 
Other settings follow Section~\ref{sec:setup}.

Figure~\ref{fig:qa1_length_sweep} shows that the main advantage emerges at long inputs. Accuracy at 0K and 1K varies by model and does not establish a consistent lead, while the proposed method reaches or ties the highest score at 16K on all three models. At 128K and longer, it remains nonzero in every measured cell. Compression reaches zero throughout this range, truncation and StreamingLLM recover only one Llama-3.1-8B sample at 2M, and Chunk RAG succeeds inconsistently. 
At 2M, the proposed method reaches 50\% on Phi-3.5-mini, 67\% on Qwen3-4B, and 80\% on Llama-3.1-8B, exceeding every plotted method. 
These observations demonstrate recall of selected facts after source removal, although the declines at the longest inputs show that bounded storage does not ensure lossless recall.

Figure~\ref{fig:qa2_length_sweep} shows a narrower advantage when an answer requires two facts. The proposed method remains nonzero on Phi-3.5-mini from 128K to 1M and on Qwen3-4B from 128K to 512K, whereas the evaluated text methods reach zero beyond their pretrained windows. Llama-3.1-8B instead falls to zero after 128K despite its stronger QA1 results. This difference shows that reliable single-fact recall does not imply successful composition. At 1M, only Phi-3.5-mini answers a question correctly, reaching 50\% on two questions. The additional Phi-3.5-mini 2M cell, outside the plotted range, contains one failed question. Table~\ref{tab:attribution} in Section~\ref{sec:ablation} attributes the QA2 failures to the routing and answer-model stages.

\subsection{Recall in Encyclopedic Prose}
This experiment measures recall in dense encyclopedic prose on RULER QA.
The per-layer allocation is $C{=}1024$ up to 1M tokens and $C{=}2048$ at 2M tokens. These limits are lower than the narrative allocation. Encyclopedic sentences carry denser factual content and cause stronger inter-anchor interference, and the lower allocation preserves reconstruction fidelity.

\begin{table}[tb]
\caption{RULER QA percentages on the three models, rounded to the nearest integer. Proposed-method sample counts appear in Table~\ref{tab:samples}. Compression and closed book each use eight questions per model and length. A dash marks a full-context cell beyond the configured evaluation limit.}
\label{tab:ruler}
\centering
\small
\setlength{\tabcolsep}{3pt}
\begin{tabular}{@{}llrrrrr@{}}
\toprule
Model & Method & 16K & 128K & 512K & 1M & 2M \\
\midrule
Phi-3.5-mini & Ours & 13 & 50 & 50 & 100 & 100 \\
 & Chunk RAG & 38 & 50 & 0 & 0 & 0 \\
 & StreamingLLM & 25 & 0 & 0 & 0 & 0 \\
 & Truncation & 25 & 0 & 0 & 0 & 0 \\
 & Full context & 63 & -- & -- & -- & -- \\
 & Compression & 38 & 25 & 0 & 0 & 0 \\
 & Closed book & 13 & 13 & 13 & 13 & 13 \\
\midrule
Qwen3-4B & Ours & 50 & 67 & 67 & 33 & 50 \\
 & Chunk RAG & 25 & 67 & 67 & 0 & 50 \\
 & StreamingLLM & 50 & 67 & 67 & 33 & 0 \\
 & Truncation & 50 & 67 & 67 & 33 & 0 \\
 & Full context & 38 & -- & -- & -- & -- \\
 & Compression & 38 & 38 & 13 & 0 & 0 \\
 & Closed book & 38 & 38 & 38 & 38 & 38 \\
\midrule
Llama-3.1-8B & Ours & 75 & 100 & 100 & 100 & 0 \\
 & Chunk RAG & 38 & 50 & 0 & 0 & 0 \\
 & StreamingLLM & 100 & 50 & 50 & 0 & 0 \\
 & Truncation & 100 & 50 & 50 & 0 & 0 \\
 & Full context & 50 & -- & -- & -- & -- \\
 & Compression & 50 & 25 & 13 & 0 & 0 \\
 & Closed book & 25 & 25 & 25 & 25 & 25 \\
\bottomrule
\end{tabular}
\end{table}

Table~\ref{tab:ruler} shows that the long-input advantage depends on the model in encyclopedic prose. The proposed method records higher scores than every compared approach on Phi-3.5-mini from 512K to 2M and on Llama-3.1-8B from 128K to 1M. Qwen3-4B instead ties several text-based methods at long inputs, including Chunk RAG at 2M, and Llama-3.1-8B fails at 2M. The Qwen3-4B score of 33\% at 1M stays lower than its 38\% closed-book score. The QA1 advantage does not extend uniformly to denser factual text. The QA1 advantage does not extend uniformly to denser factual text. The 100\% Phi-3.5-mini scores at 1M and 2M each represent one successful question, as specified in Table~\ref{tab:samples}.

Nonzero closed-book scores show that answer accuracy alone cannot establish successful document recall. Those scores use eight questions per model and are not paired estimates for the smaller long-input subsets. Section~\ref{sec:ablation} returns to these cells. Table~\ref{tab:attribution} attributes the observed failures to the routing and reconstruction stages. Table~\ref{tab:anchor_example} contrasts reconstructed content in two encyclopedic conditions.

\begin{figure*}[t]
\centering
\includegraphics[width=0.99\textwidth]{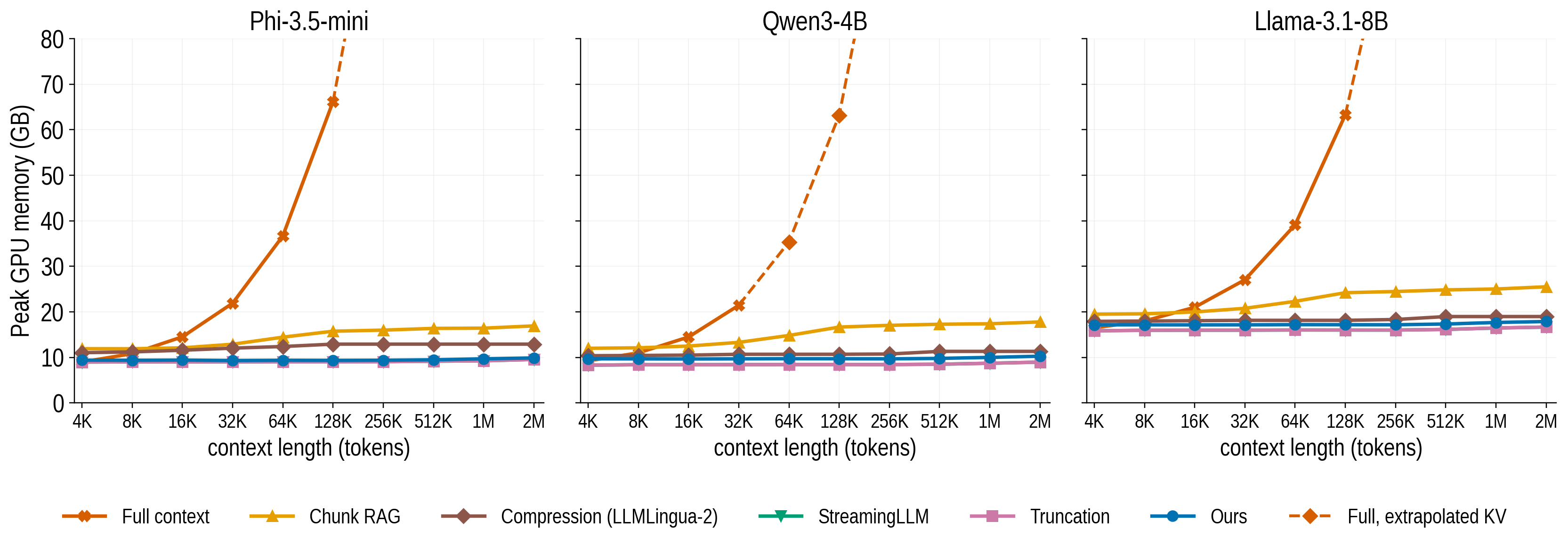}
\Description{Line charts of peak GPU memory against input length for three models. Solid full-context curves rise to each feasible endpoint, and dashed diamond continuations extrapolate KV cache growth toward the 80 GB device limit. The proposed method remains near the frozen-model's GPU memory usage.}
\caption{Measured peak GPU memory against input length. Solid full-context segments show measured runs, and dashed diamond continuations extrapolate KV cache growth toward the 80 GB device limit. The StreamingLLM series overlaps truncation.}
\label{fig:peakmem}
\end{figure*}

\subsection{Memory Scaling}
\label{sec:results-memory}

Figure~\ref{fig:peakmem} shows little change in response-time GPU memory from 4K to 2M despite the increase in source length. The two smaller models remain between 9.3 and 10.3 GB, and Llama-3.1-8B remains between 17.1 and 17.9 GB. The activation store grows by less than 0.8 GB in this range. Full attention exceeds the proposed memory profile from the 8K input length onward on every model and continues to grow as token-level state accumulates. The dashed continuations are extrapolations rather than measured runs. The measured profile reflects retention bounded by fact capacity rather than complete source length.

The bounded token approaches also limit active-context memory, but Figure~\ref{fig:qa1_length_sweep} shows that their long-input recall is lower. Chunk RAG preserves query-time access to source text at a larger measured GPU usage. Relative to truncation, it adds 2.9 to 3.7 GB at 4K and 7.4 to 8.9 GB at 2M for its reranker and chunk index. Both accuracy and memory results indicate that the main advantage of the proposed method is selective recall at nearly unchanged GPU usage.

\section{Ablation Studies}
\label{sec:ablation}

\begin{table}[tb]
\centering
\small
\setlength{\tabcolsep}{2.5pt}
\caption{Proposed-method outcomes in the reported cells. BABILong rows cover 128K to 2M. RULER 16K is separated from the longer inputs and lies inside every pretrained context window. Extr., Rout., Recon., and Ans.\ denote the recorded extraction, routing, reconstruction, and answer-model failure categories. Each question contributes to one column. Failures from cells without surviving per-question records count as unattributed, and the released data lists the per-question labels. These questions remain included in $n$ and accuracy.}
\label{tab:attribution}
\begin{tabular}{@{}lrrrrrrr@{}}
\toprule
Benchmark & Success & Extr. & Rout. & Recon. & Ans. & Unattr. & $n$ \\
\midrule
BABILong QA1 & 48 & 0 & 1 & 1 & 2 & 3 & 55 \\
BABILong QA2 & 11 & 0 & 8 & 3 & 7 & 3 & 32 \\
RULER QA 16K & 11 & 0 & 9 & 2 & 2 & 0 & 24 \\
RULER QA 128K--2M & 15 & 0 & 4 & 3 & 0 & 1 & 23 \\
\bottomrule
\end{tabular}
\end{table}

Table~\ref{tab:attribution} relates the task-level results to different failure stages. The BABILong QA2 failures involve both routing and answer-model composition, whereas classified RULER failures at longer inputs occur during routing or reconstruction. The failed RULER reconstructions carry single-word generic anchors such as ``Duchy'', and the QA2 routing misses concentrate on questions about carried objects. The distinction motivates separate analyses of task requirements, reconstruction capacity, evidence position, and implementation choices. Seven failures remain unattributed. The absence of extraction-labeled failures does not extend to that subset.

\subsection{Performance by Reasoning Requirement}

\begin{figure*}[t]
\centering
\includegraphics[width=\textwidth]{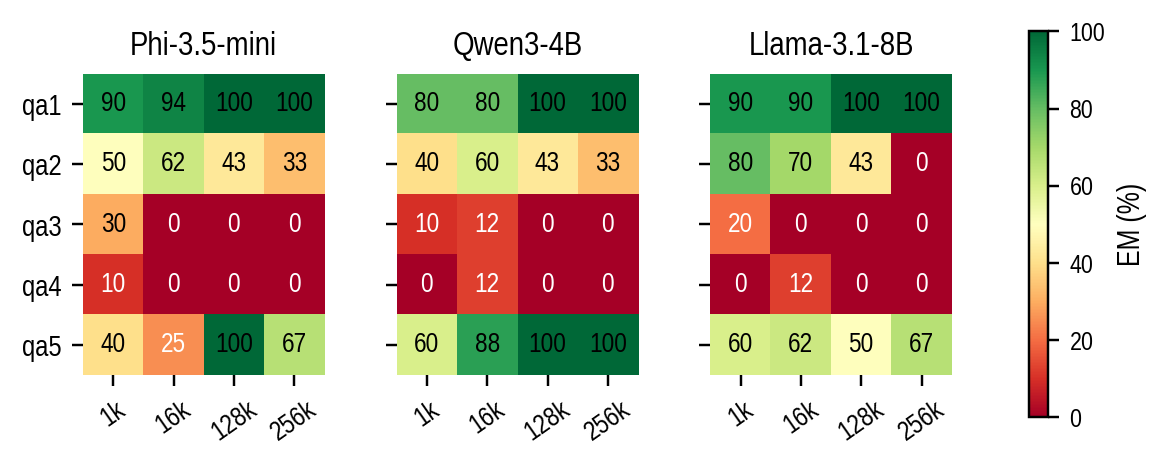}
\Description{Heatmap of exact match on five BABILong tasks and four input lengths on three models. QA1 ranges from eighty to one hundred percent and reaches one hundred percent at 128K and 256K. QA2 varies by model and length. QA3 and QA4 approach zero at longer lengths, while QA5 remains model dependent.}
\caption{The proposed method's exact match on the BABILong task and length grid. Cells pool every measured sample. At QA2 128K, each model uses seven samples and achieves 42.9\%, whereas the length sweep uses three samples per model.}
\label{fig:heatmap}
\end{figure*}

Figure~\ref{fig:heatmap} distinguishes direct recall from tasks requiring additional relations or composition. QA1 achieves 100\% at 128K and 256K on all models, while QA3 and QA4 remain at or near zero from 16K in the BABILong benchmark. Stage diagnostics distinguish these failures. QA4 loses evidence during extraction because relevant sentences can lack the entity, numeral, or identifier cues used by the filter. QA3 fails despite full extraction coverage and reconstruction fidelity of at least 0.77. Retaining evidence alone does not ensure correct composition. QA5 remains model dependent, indicating that the number of event arguments alone does not explain the task ordering.

\subsection{Reconstruction Capacity and Layer Allocation}

\begin{figure*}[tb]
\centering
\begin{subfigure}[b]{0.49\textwidth}
\centering
\includegraphics[width=\linewidth]{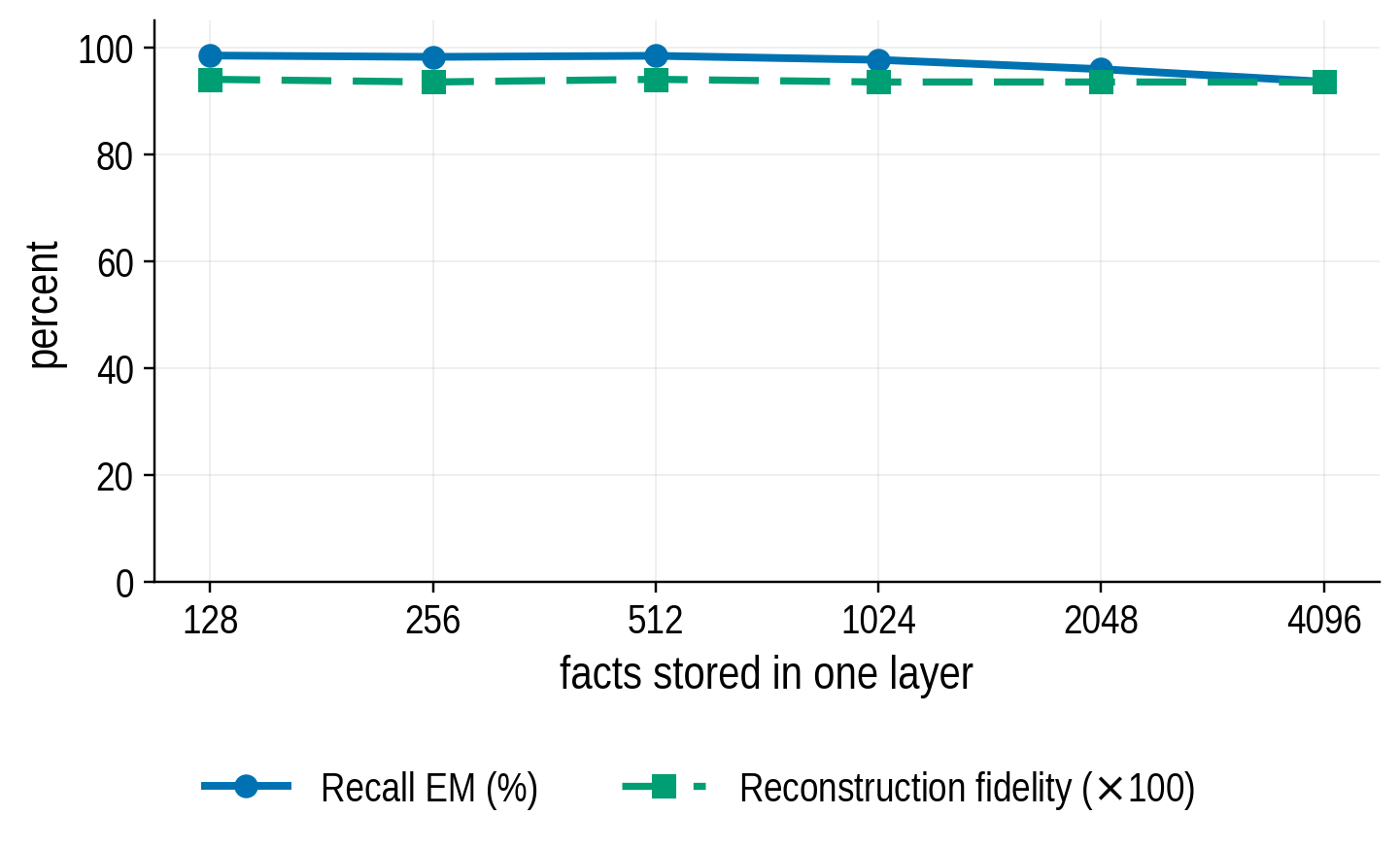}
\Description{A line chart of Phi-3.5-mini template-fact capacity at one layer. Mean recall stays above ninety-three percent and mean fidelity near zero point nine four from 128 to 4,096 facts.}
\caption{}
\label{fig:capacity_template}
\end{subfigure}
\hfill
\begin{subfigure}[b]{0.49\textwidth}
\centering
\includegraphics[width=\linewidth]{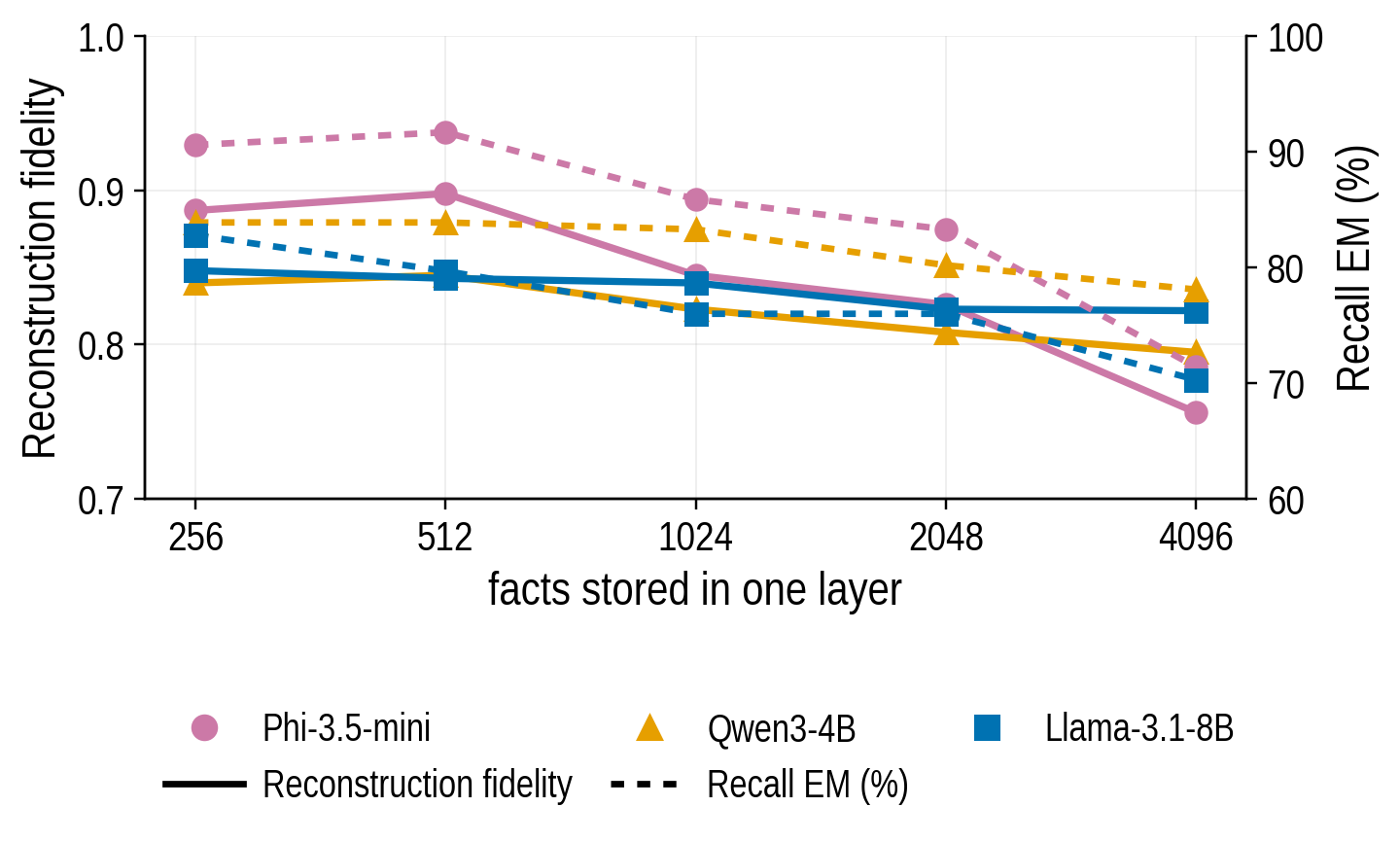}
\Description{One panel comparing three models on one shared narrative-fact pool. Solid lines show fidelity on the left axis and dashed lines show recall on the right axis. Both decline as fact count grows, and exact recall does not follow the fidelity ordering.}
\caption{}
\label{fig:capacity_narrative}
\end{subfigure}
\caption{Single-layer capacity at two fact representations. (a) is short key and numeric-code templates on Phi-3.5-mini. (b) is one shared pool of narrative sentences on three models, where solid lines show reconstruction fidelity on the left axis and dashed lines show recall EM on the right axis.}
\label{fig:capacity}
\end{figure*}

Figure~\ref{fig:capacity} shows that reconstruction quality depends on fact representation as well as store size. In Figure~\ref{fig:capacity}, Recall EM measures exact reproduction of a stored value rather than question-answering accuracy. In the template sweep of Figure~\ref{fig:capacity_template}, mean recall remains between 93.6 and 98.5\% from 128 to 4{,}096 facts, and fidelity remains near 0.94. This result shows no abrupt capacity transition for the short templates.

The shared narrative pool in Figure~\ref{fig:capacity_narrative} produces a different profile. Fidelity and exact recall decline overall as the stored fact count increases, although neither changes monotonically at every step. At 4{,}096 facts, Llama-3.1-8B has the highest fidelity, whereas Qwen3-4B has the highest recall EM. Partial token agreement and exact sentence recovery produce different model rankings. The contrast to the template sweep also indicates that a nominal allocation limit cannot guarantee the same reconstruction quality for arbitrary prose.

Table~\ref{tab:anchor_example} illustrates variation between two encyclopedic conditions. The compound anchor at 2M recovers the relevant sentence, while the generic anchor at 1M produces repetition despite twice the per-layer load. The sample-level fidelity averages all routed facts rather than only the successful sentence. This example motivates attention to anchor specificity but does not isolate its effect from input length and storage load.

\begin{table}[tb]
\centering
\caption{Reconstructions for the same SQuAD question on Qwen3-4B at two input lengths. Fid.\ reports the sample-level mean of all routed facts. The conditions differ in anchor wording and storage load.}
\label{tab:anchor_example}
\small
\setlength{\tabcolsep}{3pt}
\begin{tabular}{@{}lp{0.17\columnwidth}p{0.42\columnwidth}cc@{}}
\toprule
Cond. & Anchor & Reconstruction (excerpt) & Fid. & EM \\
\midrule
1M & ``Duchy'' & ``...was under the a of the Normans, a Norman, a Norman, a Norman, ...'' & 0.30 & 0 \\
2M & ``Duchy of Normandy'' & ``The Duchy of Normandy ... was a great fief of medieval France, and under Richard I ...'' & 0.75 & 1 \\
\bottomrule
\end{tabular}
\end{table}

Table~\ref{tab:implementation_controls} examines layer allocation on paired development items. Increasing $K$ from 1 to 6 raises accuracy from 62.5 to 87.5\% while routing recall remains at 100\%, locating the observed improvement after routing. The $K{=}12$ result comes from a different evaluation split and cannot extend this comparison into a controlled three-point trend.

\begin{figure*}[tb]
\centering
\includegraphics[width=\textwidth]{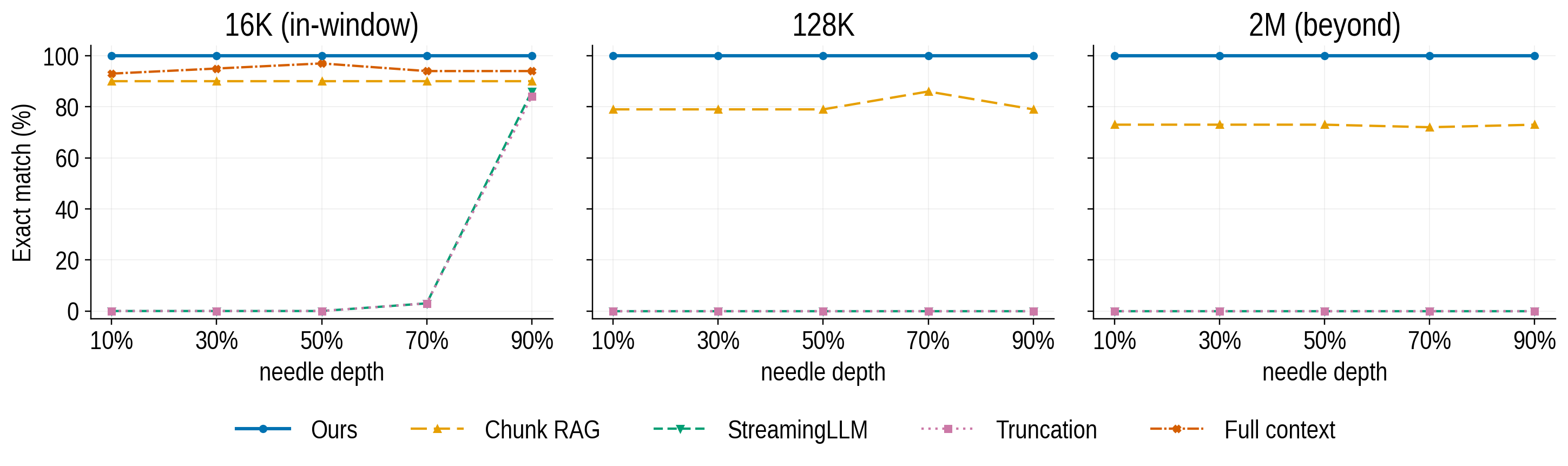}
\Description{Line charts for Phi-3.5-mini showing accuracy against needle depth at 16K, 128K, and 2M. The proposed method stays at one hundred percent. Chunk RAG ranges from seventy-two to ninety percent. Truncation and streaming succeed only near the retained tail at 16K.}
\caption{Phi-3.5-mini accuracy against needle depth on the paraphrase set at three lengths.}
\label{fig:depth}
\end{figure*}

\subsection{Sensitivity to Evidence Position}

Figure~\ref{fig:depth} examines whether recall depends on the source position of a factual sentence in the controlled paraphrase set. The proposed method reaches 100\% at all five tested positions and all three lengths, including 2M. Truncation and StreamingLLM instead concentrate their 16K successes near the source tail and fail at the longer inputs. Chunk RAG varies little by position at 2M but achieves only 72 to 73\%. In this evaluation, the proposed method's advantage combines stable recall at every tested source position and higher accuracy after the source exceeds the context window.

\begin{table*}[tb]
\centering
\caption{Measured implementation controls. Accuracy reports task EM except in the deletion rows, which report deleted-fact recall and the share of retained outputs that changed. Fid. is reconstruction fidelity. The $K{=}12$ row uses the main evaluation split and is not paired to the $K{=}1,6$ development rows. A dash denotes an inapplicable metric.}
\label{tab:implementation_controls}
\begin{tabular}{@{}lllrrr@{}}
\toprule
Study & Model & Setting & Accuracy & Fid. & $n$ \\
\midrule
Layer count & Phi-3.5-mini & $K{=}1$ & 62.5 & 0.88 & 8 \\
Layer count & Phi-3.5-mini & $K{=}6$ & 87.5 & 1.00 & 8 \\
Layer-count reference & Phi-3.5-mini & $K{=}12$ & 93.8 & 0.95 & 16 \\
Encoding steps & Qwen3-4B & 20 guided, 10 refinement & 60.0 & 0.87 & 5 \\
Encoding steps & Qwen3-4B & 60 guided, 30 refinement & 60.0 & 0.99 & 5 \\
Encoding steps & Llama-3.1-8B & 20 guided, 10 refinement & 80.0 & 1.00 & 5 \\
Gate threshold & Qwen3-4B & $\theta{=}0.45$ & 60.0 & 0.97 & 5 \\
Gate threshold & Qwen3-4B & $\theta{=}0.75$ & 60.0 & 0.99 & 5 \\
Gate threshold & Llama-3.1-8B & $\theta{=}0.45$ & 40.0 & 0.62 & 5 \\
Gate threshold & Llama-3.1-8B & $\theta{=}0.75$ & 80.0 & 1.00 & 5 \\
Semantic routing & Qwen3-4B & dense only & 60.0 & 0.99 & 5 \\
Semantic routing & Llama-3.1-8B & dense only & 80.0 & 1.00 & 5 \\
Deletion & Phi-3.5-mini & 8 selected entries removed & 0.0 & -- & 8 \\
Deletion & Qwen3-4B & 8 selected entries removed & 0.0 & -- & 8 \\
Deletion & Llama-3.1-8B & 8 selected entries removed & 0.0 & -- & 8 \\
Deletion locality & Phi-3.5-mini & retained outputs changed & 0.0 & -- & 56 \\
Deletion locality & Qwen3-4B & retained outputs changed & 0.0 & -- & 56 \\
Deletion locality & Llama-3.1-8B & retained outputs changed & 0.0 & -- & 56 \\
\bottomrule
\end{tabular}
\end{table*}

\subsection{Encoding and Deletion}

The encoding control in Table~\ref{tab:implementation_controls} separates reconstruction quality from final accuracy. On five Qwen3-4B questions, reducing the optimization schedule lowers fidelity from 0.99 to 0.87 but leaves EM at 60\%. The longer schedule improves token-level reconstruction without raising aggregate answer accuracy. The Llama-3.1-8B row evaluates only the shorter schedule and reaches fidelity 1.00, and no cross-model schedule conclusion follows from these unmatched rows.

The deletion rows test whether removal affects the remaining entries. After eight entries are deleted from a fully recalled 64-fact template store, deleted-fact recall falls to zero and all 56 retained outputs remain unchanged on each model. Removing an entry excludes its key and residual vector from Equation~\eqref{eq:read}. The matched result on the three models demonstrates local removal at the tested scale.

\begin{figure}[tb]
\centering
\includegraphics[width=0.99\columnwidth]{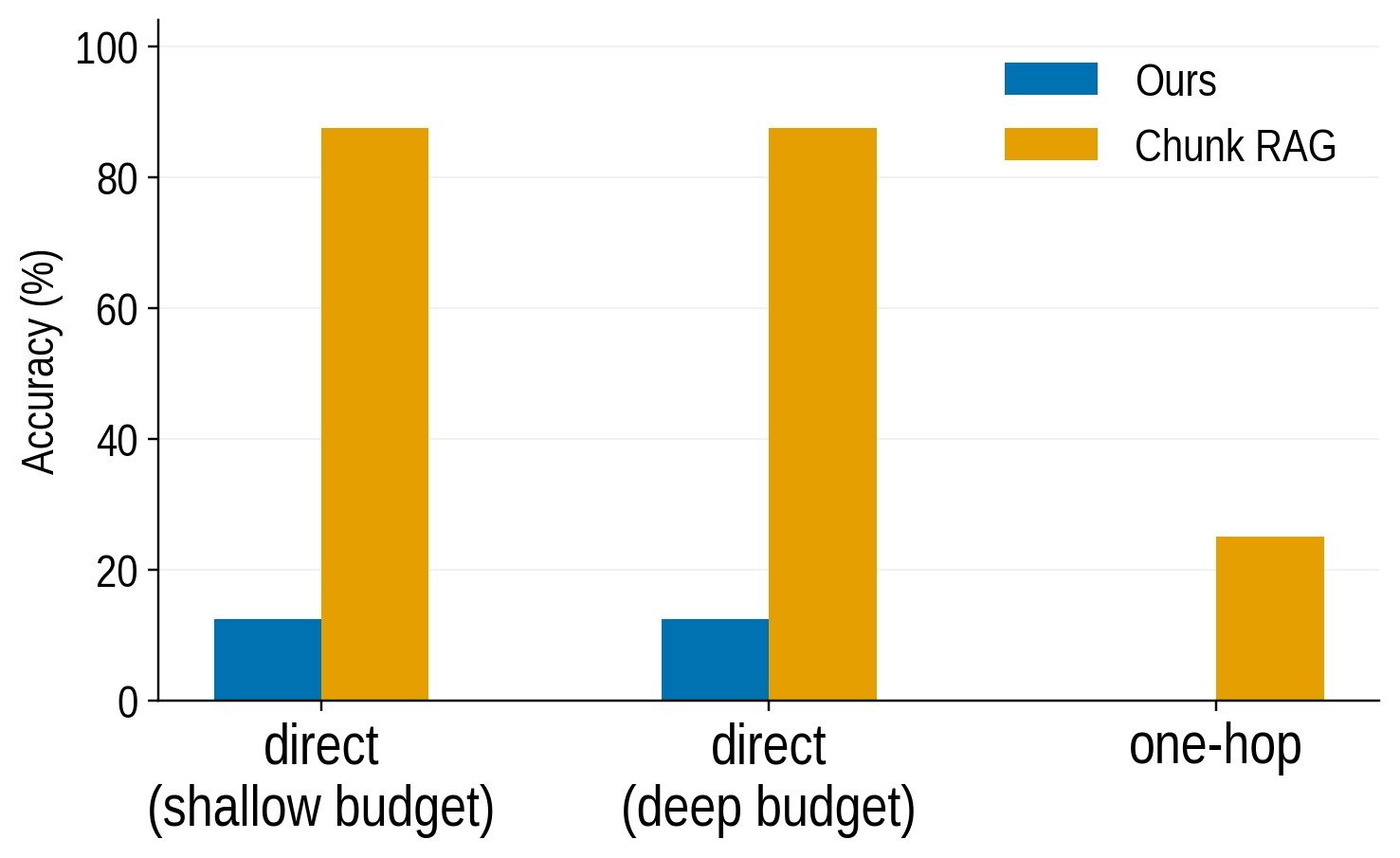}
\Description{Grouped bar chart comparing the proposed method and Chunk RAG on a routing probe at 256 thousand tokens. Direct-question accuracy is 12.5\% for the proposed method and 87.5\% for Chunk RAG at shallow and deep budgets. One-hop accuracy at the deep budget is zero and 25\%.}
\caption{Routing-stage probe at reduced lexical overlap, built from the NoLiMa needle set at 256K. Shallow and deep denote $\hat{m}{=}1$ and $\hat{m}{=}3$ for direct questions. The one-hop condition uses $\hat{m}{=}3$.}
\label{fig:routingprobe}
\end{figure}

\subsection{Routing and Gating}

Figure~\ref{fig:routingprobe} shows that increasing retention depth does not resolve the direct-question failures in the NoLiMa-derived probe. The proposed method answers one of eight questions at both budgets, compared to seven of eight for Chunk RAG. It also fails on all four one-hop questions. The recorded missing or spurious anchors identify a limitation in access to stored facts that the larger retention budget does not remove. One successful query reaches a stored character fact because ``vegan'' occurs in its value, and the Lahijan query instead matches auxiliary words to unrelated anchors from filler prose. This result qualifies the position stability in Figure~\ref{fig:depth}, because a fact can remain recallable from different source positions but stay difficult to select from a differently worded question.

The dense-only rows in Table~\ref{tab:implementation_controls} report performance after lexical ranking is removed, but the table contains no matched hybrid reference for those items. They do not establish a benefit from reciprocal-rank fusion. The gate comparison instead reveals model-dependent sensitivity. Lowering $\theta$ from 0.75 to 0.45 reduces Llama-3.1-8B accuracy from 80 to 40\% and fidelity from 1.00 to 0.62, while Qwen3-4B accuracy remains unchanged. A more permissive gate can accompany poorer reconstruction rather than improved access.

\section{Discussion and Limitations}
\label{sec:discussion}

The results identify selective factual recall as the main benefit of bounded activation memory, while multi-fact composition and encyclopedic recall vary by model and question. QA4 of the BABILong benchmark loses evidence during extraction, and QA3 fails despite full coverage and high fidelity, placing that limit in answer-model composition. The NoLiMa probe adds anchor routing as a limitation when questions have limited lexical or semantic correspondence to retained anchors. Broad synthesis, implicit relations, and unstored context remain better suited to raw-text retrieval or full-context processing when feasible. In interaction terms, these failures correspond to questions people ask naturally in a long conversation, such as how an earlier state changed, relations stated without an explicit name, and questions worded differently from the source. An assistant that misses them cannot follow the conversation fully, and closing this gap is the main direction for future work.

The main drawback is the time that the structure of the method itself requires. Memorization runs a gradient optimization for every retained fact, and its cost grows as $\mathcal{O}(N)$ optimization loops of fixed step count, far more model passes than the single embedding pass that retrieval indexing needs. At response time, routing compares the question against all $N$ stored entries, and recall then decodes every routed fact in its own generation pass before the answer pass, a per-question latency that single-pass methods avoid. These costs are structural rather than incidental, and the encoding cost amortizes only when later questions reuse the retained entries.

The evaluation covers three model families up to 8B parameters, one GPU class, and English automated benchmarks. The one-million-token and two-million-token cells contain one to five samples, and their scores indicate feasibility rather than precise accuracy estimates. The full-context reference stops at the 40{,}000-token harness limit, and the StreamingLLM approach implements a prompt retention proxy. Reconstruction is deterministic at the algorithm level, but the evaluation does not establish bitwise reproducibility between hardware or numerical configurations. The benchmarks are automated and collect no human feedback, and trust in reconstructed answers, recovery after a recall failure, and longitudinal use remain unmeasured.

\FloatBarrier
\section{Conclusion}

This paper proposes a method for recalling facts from past long contexts that exceed a language model's context window without the need for model-weight editing or additional training.
The proposed method assigns the facts of a long context to feed-forward layers of the LLM, stores the encoded values as residual vectors, and at answer time recalls past facts by reconstructing them from the stored residual vectors.
At two million tokens, the proposed method still answers at least half of the single-fact questions for all three models, a length where every compared training-free method falls to zero or near zero.
Its recall remains unchanged at every tested location of the fact in the source, GPU memory at response time remains close to that of the frozen model alone, and stored entries occupy less space than the KV cache they replace.
Moreover, repeated access to the same entry returns the same fact, and deleting an entry leaves the answers from the remaining facts unchanged in the deletion test.
This shows that the proposed method can provide reproducible recall and selective fact removal in scenarios where users and agents converse and interact for long periods. Future work will broaden relational coverage, strengthen semantic routing, and evaluate privacy and interactive use.

\section*{Generative AI Disclosure}
Anthropic Claude assisted in implementing the research ideas in code. Anthropic Claude and OpenAI Codex assisted in revising the author-written manuscript. All AI-assisted content was reviewed by the authors, who remain responsible for the work.

\section*{Ethics and Privacy Statement}
The proposed method stores facts extracted from user-provided documents. Deployed memory can consequently retain sensitive statements after source disposal. In the 64-fact deletion test, removing 8 entries leaves model weights fixed and changes none of the 56 retained outputs. The evaluation uses public benchmark corpora and involves no human subjects. Deployment requires access control for the stored memory and its routing index, provenance tracking, and policies for updating or deleting stale information. Adversarial routing and privacy leakage remain untested deployment risks.

\bibliographystyle{ACM-Reference-Format}
\bibliography{refs}

\end{document}